\PassOptionsToPackage{dvipsnames,table}{xcolor}  
\documentclass{article}

\PassOptionsToPackage{numbers}{natbib}  

 \usepackage[main, final]{neurips_2026}

\usepackage[utf8]{inputenc} 
\usepackage[T1]{fontenc}    
\usepackage[pagebackref,breaklinks,colorlinks,citecolor=blue, linkcolor=blue]{hyperref}
\usepackage{url}            
\usepackage{booktabs}       
\usepackage{amsfonts}       
\usepackage{nicefrac}       
\usepackage{microtype}      
\usepackage{xcolor}         

\usepackage{graphicx}
\usepackage{subcaption}
\definecolor{softgreen}{RGB}{34,139,34}
\definecolor{softgray}{RGB}{125,125,125} 
\usepackage{tikz}
\usepackage{pgfplots}
\usepackage{xspace}
\usepackage{amsmath}

\usepackage[table]{xcolor}
\usepackage{times}
\usetikzlibrary{positioning,fit,calc}
\newcommand{\ours}{\texttt{V-GIFT}\xspace}

\definecolor{baselinecolor}{rgb}{0.9, 0.9, 1.0}
\usepackage{makecell}
\newcommand{\parag}[1]{\smallskip\noindent\textbf{#1}\enspace}

\usepackage{orcidlink}
\usepackage{multirow}       
\usepackage{svg}
\usepackage{caption}
\usepackage{wrapfig} %

\title{\ours: Boosting Visual Instruction Tuning with\\ Self-Supervised Guidance}

\author{%
  Sophia Sirko-Galouchenko$^{1,2}$ \quad
  Monika Wysocza\'nska$^{1}$ \quad
  Andrei Bursuc$^{1}$ \\
  \textbf{Nicolas Thome$^{4,3,2}$ \quad Spyros Gidaris$^{1}$} \\[4pt]
  $^{1}$Valeo.ai \quad
  $^{2}$Sorbonne Universit\'e, CNRS, ISIR, F-75005 Paris, France \\
  $^{3}$Institut universitaire de France (IUF) \quad $^{4}$ ILLS, CNRS, Montreal, QC H2S 3H1
}

\begin{document}

\maketitle

\begin{abstract}
Multimodal large language models (MLLMs) perform well on many vision-language tasks but often struggle with vision-centric problems that require fine-grained visual perception. Recent evidence suggests that this limitation arises not from weak visual representations, but from under-utilization of visual information during instruction tuning, where many tasks can be partially solved using language priors alone. 
We propose a simple and lightweight approach that augments visual instruction tuning with a small number of visually grounded self-supervised tasks expressed as natural language instructions. By reformulating classical self-supervised pretext tasks, such as rotation prediction, color matching, and cross-view correspondence, as image–instruction–response triplets, we introduce supervision that cannot be solved without relying on visual evidence. Our approach requires no human annotations, no architectural modifications, and no additional training stages.
Across multiple models, training regimes, and benchmarks, injecting only a small fraction (3–10\%) of such visually grounded instructions consistently improves performance on vision-centric evaluations. Our findings highlight instruction tuning with visually grounded SSL tasks as a powerful lever for improving visual perception in MLLMs through simple adjustments to the training data distribution. \href{https://github.com/sirkosophia/V-GIFT}{Project page}
\end{abstract}
\begin{center}
  \centering
  \includegraphics[width=\textwidth]{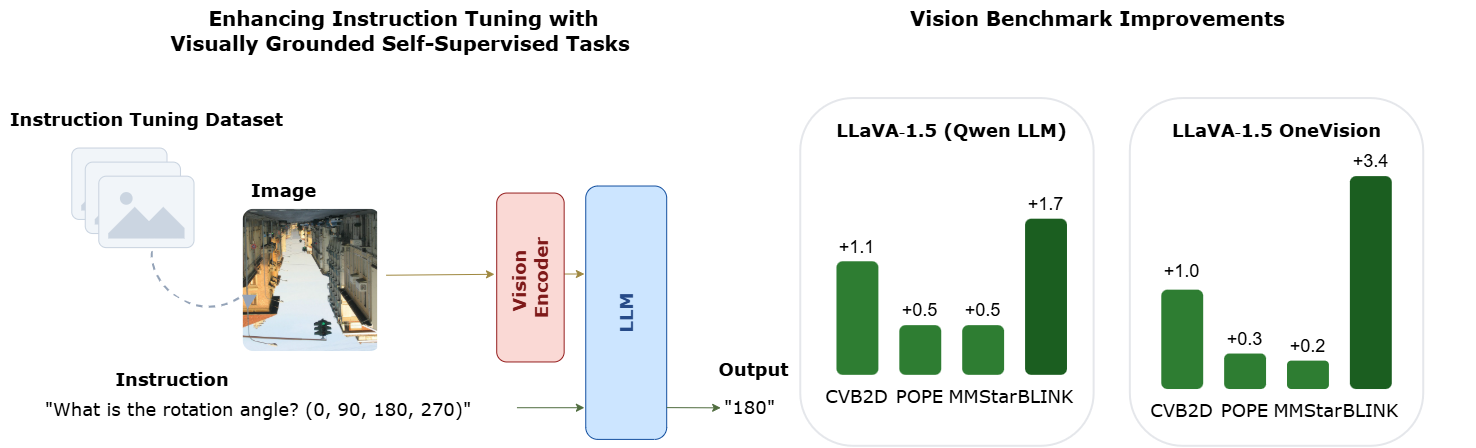}
  \captionof{figure}{\textbf{V}isually \textbf{G}rounded \textbf{I}nstruction \textbf{F}ine-\textbf{T}uning \ours. We enhance visual instruction tuning by injecting visually grounded self-supervised tasks as additional instruction-following examples sampled from the instruction-tuning data (left; rotation prediction shown). This simple modification encourages better use of visual information and yields consistent gains on vision-centric benchmarks (right; CVB-2D, POPE, MMStar, BLINK) across model variants.}
  \label{fig:teaser}
\end{center}
  
\section{Introduction}
\label{sec:intro}

Multimodal Large Language Models (MLLMs)~\citep{alayrac2022flamingo, li2023blip, liu2024improved} combine pretrained vision encoders with large language models (LLMs)~\citep{brown2020language, touvron2023llama, team2024qwen2} to perform multimodal instruction following.  
Modern open-source systems~\citep{liu2023visual, liu2024improved, li2024llava, lin2024vila} typically consist of three components: a pretrained vision encoder, a lightweight projection module that maps visual features into the language embedding space, and a pretrained LLM decoder. Trained via vision–language alignment followed by visual instruction tuning, these models achieve strong performance on image captioning, visual question answering, and general multimodal dialogue.

Despite their apparent competence, MLLMs frequently stumble on \emph{vision-centric tasks} that require fine-grained visual understanding, such as object counting, spatial positioning, and geometric relation understanding \citep{fu2024blink, tong2024cambrian, tong2024eyes}. While one might expect these failures to stem from weak visual representations~\citep{tong2024eyes}, recent evidence suggests a more nuanced bottleneck: modern encoders such as CLIP~\citep{radford2021learning} and DINOv2~\citep{oquab2023dinov2} already capture rich, task-relevant features, yet the LLM component often under-utilizes this visual information during decoding \citep{fu2025hidden}. This shift in understanding reframes the problem: the issue is not merely representational capacity, but training dynamics. During visual instruction tuning, models are optimized on image–instruction–response triplets expressed entirely in natural language. While flexible, this formulation introduces an unintended bias: many instruction-tuning examples can be partially or fully solved using strong language priors alone~\citep{deng2025words}. Besides, most of the times, these examples are automatically generated from generic captions of the images~\citep{liu2023visual} without enforcing the use of vision cues for solving them. As a result, the model may learn language-dominant shortcut strategies and under-rely on visual evidence, even when vision is necessary. Notably, this limitation persists even as models scale and training data increases \citep{deng2025words}.

We hypothesize that this imbalance in supervision during instruction tuning is a central cause of weak vision-centric perception. If models are rarely required to depend on visual input in order to succeed, they tend to default to language-based heuristics.
We propose to address this supervision imbalance within the visual instruction tuning phase. Our key idea is to reinterpret instruction tuning as a modality competition process: when many tasks can be solved through linguistic priors alone, training may implicitly favor language-dominant strategies. To counteract this bias, we inject inherently 
vision-forcing tasks into the instruction distribution.
Specifically, we reformulate self-supervised (SSL) tasks, such as rotation prediction~\citep{gidaris2018unsupervised}, color matching~\citep{zhang2016colorful}, and point correspondence~\citep{wang2019learning}, as image–instruction–response triplets compatible with standard MLLM pipelines. These tasks possess two critical properties:  
\textbf{(1)} they are visually grounded by construction, the correct answer cannot be inferred from language priors alone; and  
\textbf{(2)} they require no human annotation, as supervision is derived automatically from image transformations or feature-based correspondences.

Unlike recent works that incorporate self-supervised objectives through auxiliary losses or reinforcement learning with verifiable rewards (RLVR), our method does not modify the optimization paradigm. We retain standard autoregressive cross-entropy training and instead adjust the \emph{distribution of instruction-following tasks} to systematically include vision-only problems that compel the model to rely on visual tokens, as language priors provide no information about a grayscale pixel’s original color or a randomly rotated image’s orientation. 
By integrating these tasks into visual instruction tuning, we encourage more effective coordination between visual perception and language reasoning
without requiring auxiliary losses, architectural changes, or expensive RLVR pipelines.

Across multiple benchmarks, MLLM backbones, and training regimes, we observe consistent improvements in vision-centric perception. Remarkably, injecting a small fraction (between 3\% and 10\%) of visually grounded SSL instructions yields measurable gains across diverse benchmarks, and these improvements generalize to stronger models such as OneVision-1.5~\citep{an2025llava}. Control experiments with matched training iterations confirm that the gains are not attributable to additional compute. We further show that SSL supervision is most effective when mixed directly into instruction tuning rather than applied as a separate pre- or post-training stage, highlighting the importance of shaping supervision during multimodal alignment. 
Finally, we find that even SSL tasks generated from a single image (through random view sampling)~\citep{asano2020critical} can produce improvements, suggesting that the key factor is not dataset scale but the presence of objectives that compel visual grounding.

To summarize, our contributions are:
\begin{itemize}
    \item We propose a simple yet effective framework to reformulate classic self-supervised pretext tasks as instruction-following data, mitigating language shortcuts in MLLMs and encouraging the LLM to better utilize visual representations.
    \item Our framework integrates seamlessly into the instruction-following pipeline. It is applicable to any MLLM architecture without architectural modifications and avoids additional pre- or post-training steps as well as complex hyperparameter tuning.
    \item Through extensive experiments, we show consistent improvements across models, training regimes, and benchmarks, while requiring minimal additional compute.
\end{itemize} 

\section{Related Work}
\label{sec:related_work}

\parag{Multimodal Large Language Models.}
MLLMs have emerged as a natural extension of LLMs to modalities beyond text~\citep{caffagni2024revolution}, integrating non-textual information through two dominant architectural paradigms. In cross-attention-based models~\citep{alayrac2022flamingo, laurenccon2023obelics}, visual features are injected via dedicated cross-attention layers interleaved within the LLM. In projection-based models, exemplified by LLaVA~\citep{liu2023visual, liu2024improved}, a vision encoder, typically a CLIP-style~\citep{radford2021learning, tschannen2025siglip} encoder, maps images into visual embeddings with a lightweight adapter.  
The latter has become the dominant paradigm, with most modern MLLMs adopting this design for its simplicity and ease of training~\citep{liu2024improved, li2024llava, an2025llava, bai2025qwen3,wang2025internvl3, tong2024cambrian, steiner2024paligemma,kamath2025gemma}. LLaVA models follow a two-stage recipe: adapter pretraining on image-caption pairs, followed by visual instruction tuning of the full model. Recent improvements on the data front include introducing a mid training stage~\citep{an2025llava} to inject additional knowledge before instruction tuning; or curating high-quality human-annotated visually grounded instructions yielding richer supervision and stronger spatial grounding ~\citep{deitke2025molmo}.
In this work we also take the data-centric approach to improve capabilities of MLLMs, yet we do not use any manual labels and leverage instead self-supervised pretext objectives.

\parag{Vision-centric Strategies for MLLMs.}
Improving visual perception in MLLMs focuses on two main axes. First line of work focuses on designing more expressive projectors \citep{mckinzie2024mm1, liu2024improved, cha2024honeybee}, for example by aggregating multi-layer features from the vision encoder before being fed into the LLM \citep{chen2024lion, lin2025multi}, while other studies explore the use of multiple vision encoders \citep{tong2024eyes, kar2024brave, tong2024cambrian, azadani2025leo, shi2024eagle, lu2024deepseek}.
Second line of work identifies visual bottleneck in utilization of visual information during LLM decoding \citep{fu2025hidden}. Motivated by this insight, recent works introduce auxiliary objectives that directly supervise visual tokens within the LLM decoder. These include reconstruction-based losses applied to visual token outputs \citep{wang2025reconstructive} and distillation of intermediate LLM features from external vision foundation models \citep{yoon2025visual, caffagni2025seeing}, such as DINOv2~\citep{oquab2023dinov2}. 
In contrast to these approaches, which modify model architectures or introduce auxiliary optimization objectives, we focus on instruction tuning and show that adjusting the supervision distribution through visually grounded self-supervised instructions is sufficient to improve visual information usage. As a result, our approach remains simple and lightweight: it preserves the standard MLLM training objective and architecture, with only marginal overhead from a small fraction of SSL-style instruction examples.

\parag{Leveraging Self-supervised Learning.}
Self-supervised learning (SSL) learns visual representations from unlabeled data through annotation-free pretext tasks ranging from low-level pretext tasks such as: predicting rotation angles~\citep{gidaris2018unsupervised}, relative patch positions~\citep{doersch2015unsupervised}, colorization~\citep{zhang2016colorful}, to richer, high-level objectives including contrastive learning~\citep{he2020momentum, wu2018unsupervised, chen2020improved, chen2020simple}, prototype-based clustering~\citep{caron2020unsupervised, gidaris2020learning}, self-distillation~\citep{grill2020bootstrap, caron2021emerging, oquab2023dinov2, gidaris2024moca, gidaris2021obow, venkataramanan2025franca}, and masked image modeling~\citep{he2022masked}. 
Recently, frameworks for improving visual utilization in MLLMs also build on SSL, e.g. masked image modeling~\citep{wang2025reconstructive}, adapting jigsaw solving as an RLVR post-training~\citep{wu2026visual, wang2025jigsawr1, guo2025ssl4rl, liu2025spatial}. In this work, we similarly draw inspiration from SSL pretext tasks and highlight a visually grounded data imbalance in visual instruction tuning, which we address by directly incorporating self-supervised tasks into the existing instruction format, without introducing auxiliary losses, additional training stages, or costly RLVR pipelines.

\section{Method}
\begin{figure}[t] 
\centering 
\includegraphics[width=\linewidth]{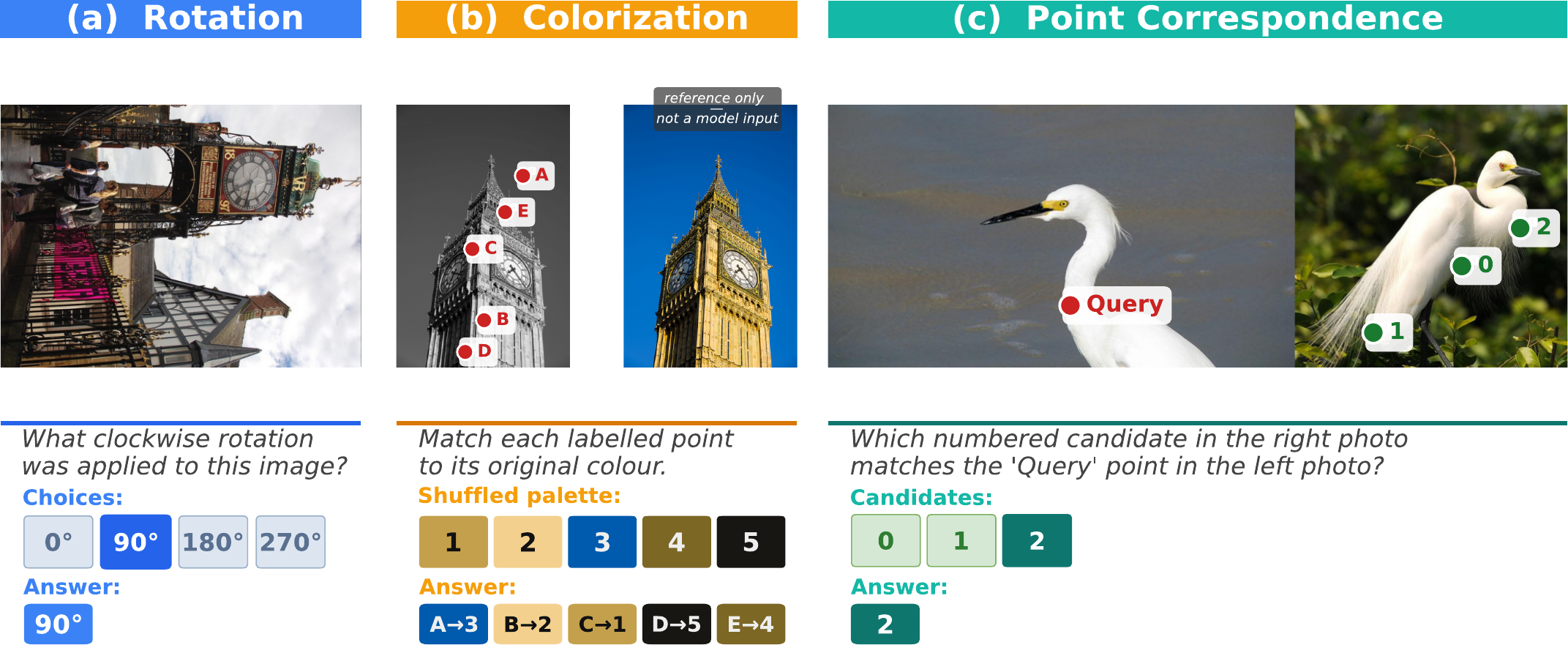} 
\caption{Visually grounded instruction-following tasks reformulated from self-supervised tasks. (a) \textbf{Rotation prediction:} the model must infer orientation from the image. (b) \textbf{Point-wise colorization:} the model must match grayscale points to their original colors, requiring fine-grained visual discrimination and spatial grounding. (c) \textbf{Point correspondence:} the model must identify corresponding points across views, requiring spatial reasoning. Overall, these tasks encourage reliance on visual evidence and integration of local and global visual cues, rather than language priors.}
\label{fig:ssl_instruct_examples}
\vspace{-10pt}

\end{figure}

Our goal is to improve the performance of MLLMs on vision-heavy tasks by encouraging stronger reliance on visual information during training. Rather than modifying model architectures or introducing additional training stages, we intervene directly in the visual instruction tuning phase. Specifically, we augment standard instruction tuning with a small number of automatically-generated visually grounded tasks, formulated as natural language instructions, that require genuine visual reasoning and cannot be reliably solved using language priors alone. 

\subsection{Multimodal Large Language Model Training Pipeline}  
\label{sec:pipeline}

We adopt a standard MLLM architecture consisting of a pretrained vision encoder, a projector, and a pretrained LLM decoder. The vision encoder extracts visual features from the input image, which are mapped by the projector into the LLM’s token embedding space and processed jointly with text tokens by the LLM.
Training follows the commonly used two-stage pipeline employed by LLaVA-style models~\citep{liu2023visual, liu2024improved, li2024llava, an2025llava}: vision–language alignment pretraining followed by visual instruction tuning.

\parag{Vision–language alignment.}  
In the first stage, the vision encoder is frozen and the projection module is trained to align visual features with the LLM embedding space using large-scale image–caption pairs. Each training sample consists of an image $I$ and an associated caption $T = (t_1, \dots, t_N)$. The model is optimized with an autoregressive language modeling objective:  
\begin{equation}  
\mathcal{L}_{\text{align}} = - \sum_{i=1}^{N} \log p_\theta(t_i \mid t_{<i}, I).  
\end{equation}  
This stage yields a base MLLM capable of jointly processing visual and textual inputs. Recently, OneVision-1.5~\citep{an2025llava} extends the vision–language alignment stage with extensive mid-training (stage 1.5) of the entire model on high-quality image–text pair data. 

\parag{Visual instruction tuning.}  
In the second stage, either the full model (e.g., OneVision-1.5~\citep{li2024llava, an2025llava}) or only the projector with the LLM (e.g., LLaVA~\citep{liu2023visual, liu2024improved}) are fine-tuned on multimodal instruction-following data. Each sample consists of an image $I$, a textual instruction $x = (x_1, \cdots, x_N)$, and a response $y = (y_1, \dots, y_M)$. The model is trained to generate the response conditioned on both the instruction and the visual input using the standard autoregressive loss:  
\begin{equation}  \label{eq:loss_instruct}
\mathcal{L}_{\text{inst}} = - \sum_{j=1}^{M} \log p_\theta(y_j \mid y_{<j}, x, I).  
\end{equation}  
This stage shapes the model’s instruction-following behavior and multimodal reasoning. \ours~ operates exclusively within this instruction tuning phase.

\subsection{Visual Self-Supervised Tasks as Instructions}  
\label{sec:ssl_tasks}

We introduce a set of visual self-supervised learning (SSL) tasks that derive supervision directly from image structure. Unlike conventional SSL methods, which rely on auxiliary losses or specialized heads during vision encoder pretraining, we reformulate these tasks as natural language instructions compatible with standard instruction tuning. This enables seamless integration into existing MLLM pipelines without architectural or optimization changes.

Each SSL task is expressed as an image–instruction–response triplet $(I, x, y)$, where the instruction $x$ specifies a visually grounded task and the response $y$ is an 
answer automatically derived from the image. These tasks follow the same format as standard multimodal instruction data and are optimized using the autoregressive cross-entropy loss in Eq.~\ref{eq:loss_instruct}.
We consider three SSL tasks (\autoref{fig:ssl_instruct_examples}).

\parag{Rotation prediction.}  
Given an image $I$, we generate a rotated version $\tilde{I} = R_{\theta}(I)$, where $\theta \in \{0^\circ, 90^\circ, 180^\circ, 270^\circ\}$ is sampled uniformly at random and independently of the image content. The instruction asks the model to identify the applied rotation and to answer only with the degree value. The response is the discrete label $y=\theta$, serialized as text. Each training example takes the form:  
\begin{equation} 
(\tilde{I},\ x=\text{``What is the rotation angle of this image?''},\ y=\text{``\{}\theta\text{\}''}).  
\end{equation}
We illustrate this rotation prediction instruction task in~\autoref{fig:ssl_instruct_examples} (a).
Because the rotation label is independent of both the instruction and dataset statistics, it cannot be predicted without inspecting the visual input. Solving the task requires inferring object orientation from visual evidence in the image.

\parag{Point-wise colorization (color matching).}  
Given a color image $I \in \mathbb{R}^{H \times W \times 3}$, we convert it to grayscale and sample $K$ spatial points $\{q_i\}_{i=1}^K$. 
For each point, we compute its average RGB color $c_i$ over a local $r \times r$ neighborhood and enforce diversity via $\|c_i - c_j\|_2 \ge \delta$, using rejection sampling, where $\delta$ is a fixed threshold. 
Each point is assigned a label (e.g., $A,B,\dots$) and overlaid on the grayscale image $\tilde{I}$.

We randomly permute the candidate colors and present them as a numbered list (RGB triplets with optional coarse color names). The instruction asks the model to match each labeled point to its corresponding color index, with responses formatted as label–index pairs:
\begin{equation}
(\tilde{I},\ x=\text{color-matching instruction},\ y=\text{``}A\text{-\{}y_A\text{\}},B\text{-\{}y_B\text{\}},\dots\text{''}),  
\end{equation}  
where $y_A$ is the index of the true color for point A. 
 
We illustrate this task in~\autoref{fig:ssl_instruct_examples} (b).
Although candidate colors are specified textually, their ordering is randomized and the correct point–color mapping is determined solely by the image. As a result, the task cannot be resolved from language priors alone. The model must localize each marked point and infer its color from visual evidence in the grayscale image, requiring fine-grained spatial grounding rather than semantic guessing.

\parag{Point correspondence.}
Given an image pair $(I_1, I_2)$ depicting the same object instance, we ask the model to identify corresponding points across views. Following DIP~\citep{sirko2025dip}, we automatically generate supervision using pseudo-segmentation masks obtained with Stable Diffusion~\citep{ssd1b} and dense DINOv2~\citep{oquab2023dinov2} features. These signals restrict point sampling to object regions with consistent semantic identity across views and enable the identification of reliable correspondences.

Specifically, we sample a query point $q$ in $I_1$, identify its best-matching location $q^{+}$ in $I_2$ via dense feature similarity, and sample two distractor points from the same object region. The candidates are randomly permuted and labeled $(0,1,2)$. The instruction asks which candidate corresponds to the query point, and the response is the index of the correct match:  
\begin{equation}
((I_1,I_2),\ x=\text{correspondence instruction},\ y=\text{``\{}y_{q^{+}}\text{\}''}),
\end{equation}  
where $y_{q^{+}} \in (0, 1, 2)$ denotes the index of the best-matching location $q^{+}$ in $I_2$ among the three candidates. 
We illustrate this task in~\autoref{fig:ssl_instruct_examples} (c). Solving it requires identifying consistent visual features across viewpoints and reasoning about spatial correspondences between the two images.
Note that our goal is not to propose a standalone self-supervised correspondence method, but to provide a simple automatically generated task for instruction tuning. We use pretrained vision models (e.g., Stable Diffusion masks and DINOv2 features) only offline to construct reliable correspondence signals from unlabeled images. These models are not involved during MLLM training, and we do not use feature distillation or auxiliary supervision.

Collectively, these tasks require the model to develop core visual reasoning abilities, including sensitivity to object geometry and orientation, fine-grained visual discrimination, precise spatial grounding, and cross-view correspondence. Solving them also requires integrating local visual cues with global image context while relying on visual evidence rather than language priors.

\subsection{Integrating SSL Instructions into Instruction Tuning}
\label{sec:integration}

We incorporate the proposed SSL tasks into visual instruction tuning by augmenting the original dataset with automatically generated SSL instruction samples. 
Let $\mathcal{D}_{\text{inst}}$ denote the original instruction dataset and $\mathcal{D}_{\text{ssl}}$ the SSL samples. The final training set is $\mathcal{D} = \mathcal{D}_{\text{inst}} \cup \mathcal{D}_{\text{ssl}}$, from which mini-batches are sampled uniformly.

We control the amount of SSL supervision via the ratio $\rho = 100 \times \frac{|\mathcal{D}_{\text{ssl}}|}{|\mathcal{D}_{\text{inst}}|}$.
which defines the percentage of additional SSL samples relative to the original dataset. 
We treat $\rho$ as a hyperparameter controlling the strength of visually grounded supervision. SSL samples can be generated from any of the tasks in Section~\ref{sec:ssl_tasks}, either individually or combined, providing complementary visual reasoning signals.

In practice, we use small ratios: $\rho=10\%$ for LLaVA-1.5 and $\rho=3\%$ for OneVision-1.5, reflecting differences in dataset scale and training dynamics. 
Despite their small proportion, these SSL samples consistently improve reliance on visual information and reduce language-driven shortcuts. 
Importantly, the additional computational cost scales linearly with $\rho$ and remains negligible in all settings, requiring no changes to the optimization schedule, batch size, or learning rate.
\section{Experiments}

\subsection{Experimental Setup}
\label{sec:4_exset}
\label{sec:experimental_setup}

\parag{Models and training protocol.}
We build on the LLaVA-1.5 framework~\citep{liu2024improved}, using both Vicuna-7B-v1.5~\citep{chiang2023vicuna} and Qwen2.5-7B~\citep{qwen2024qwen2} language model backbones and CLIP ViT-L/14 vision~\citep{radford2021learning} encoder. We evaluate two training regimes: full model fine-tuning and parameter-efficient adaptation using LoRA~\citep{hu2022lora}. To assess generalization  beyond a single architecture and training recipe, we evaluate our method on the more recent OneVision-1.5 model~\citep{an2025llava}. 
Despite the related naming, OneVision-1.5 should be viewed as a distinct model family from LLaVA-1.5: it uses different architectural components, including a RICE-ViT vision encoder~\citep{xie2025region} and a Qwen3-4B language model~\citep{yang2025qwen3}, as well as a different training pipeline, image resolution, and instruction-tuning data. 
We train the visual instruction tuning stage on LLaVA-NeXT-780k, a subset of the OneVision-1.5-Instruct-Data~\citep{an2025llava}, chosen for computational efficiency. 
We focus on these models as they provide open training recipes and instruction-tuning data, enabling modification of the instruction distribution. Other open-weight MLLMs (e.g., Qwen-VL~\citep{bai2023qwen}) do not release such data.

We augment instruction tuning with three SSL tasks (rotation, correspondence, colorization). Unless stated otherwise, $\rho=10\%$ for LLaVA-1.5 (Vicuna, Qwen) and $\rho=3\%$ for OneVision-1.5.

\parag{Training details.}
All experiments are trained on $4\times$H100 GPUs. Full fine-tuning on LLaVA-v1.5-mix665k of LLaVA-1.5-Qwen7B takes approximately 12 hours under this setup. Full fine-tuning on 110\% of visual instruction following data of LLaVA-1.5-Qwen7B requires approximately 14 hours. 
We follow the original LLaVA-1.5 and OneVision-1.5 training recipes and keep all optimization hyperparameters identical to the standard instruction tuning configuration. For the main results in \autoref{tab:main_results}, we report mean performance and standard deviation over three independent training runs with different random seeds.

\parag{Evaluation benchmarks.}
We evaluate performance on several vision-centric multimodal benchmarks: CV-Bench 2D~\citep{tong2024cambrian} (CVB-2D), POPE~\citep{li2023evaluating}, MMStar~\citep{chen2024we}, and BLINK~\citep{fu2024blink}, We additionally report results on generalist benchmarks MathVision \citep{wang2024measuring}, OCRBench \citep{liu2024ocrbench} and RealWorldQA \citep{xai2024grok}.

\subsection{Main Results on Vision-Centric Benchmarks}
\label{sec:4_main_res}
\label{sec:main_results}

As presented in \autoref{tab:main_results} across almost all evaluation settings, incorporating SSL instruction tuning consistently improves performance over the baseline models. Gains are observed for both LLaVA-1.5-Vicuna and LLaVA-1.5-Qwen, which share the same vision encoder and training pipeline but differ in the LLM backbone, indicating that the effect is not specific to the decoder. Importantly, improvements also extend to OneVision-1.5, a stronger model with a distinct architecture, pipeline, and training data, demonstrating that our method generalizes beyond a single implementation and remains effective for more recent, higher-performing MLLMs.

\begin{table}[t]
\centering
\small
\setlength{\tabcolsep}{4pt}
\caption{Main results of incorporating visually grounded SSL tasks on three MLLM models (LLaVA-1.5-Vicuna-7B, LLaVA-1.5-Qwen2.5-7B, OneVision-1.5-4B). Performance is reported on vision-centric and general benchmarks. Results are averaged over three independent training runs, with standard deviations reported as ±. Numbers in parentheses indicate improvement over the baseline.}
\resizebox{\textwidth}{!}{%
\begin{tabular}{lccccc@{\hspace{10pt}}ccc}
\toprule
& \multicolumn{5}{c}{\textbf{Vision-Centric}} & \multicolumn{3}{c}{\textbf{General}} \\
\cmidrule(lr){2-6} \cmidrule(lr){7-9}
\textbf{Method} & \textbf{CVB-2D} & \textbf{POPE} & \textbf{MMStar} & \textbf{BLINK} & \textbf{Avg.} & \textbf{MathVision} & \textbf{OCRB} & \textbf{RealWQA} \\
\midrule
\multicolumn{9}{l}{\textit{\small LLaVA-1.5-Vicuna-7B}} \\[-2pt]
Baseline & 55.9 & 87.0 & 33.5 & \textbf{38.2} & 53.6 & 12.9 & 319.3 & 54.2 \\
\rowcolor{baselinecolor}
\ours
  & \textbf{58.5} \textcolor{softgreen}{(+2.6)}
  & \textbf{87.2} \textcolor{softgreen}{(+0.2)}
  & \textbf{34.6} \textcolor{softgreen}{(+1.1)}
  & 37.8 \textcolor{softgray}{(-0.4)}
  & \textbf{54.5} \textcolor{softgreen}{(+0.9)}
  & \textbf{13.3} \textcolor{softgreen}{(+0.4)}
  & 318.3 \textcolor{softgray}{(-1.0)}
  & \textbf{55.4} \textcolor{softgreen}{(+1.2)} \\
\rowcolor{baselinecolor}
  & \scriptsize$\pm$1.52 & \scriptsize$\pm$0.29 & \scriptsize$\pm$0.25 & \scriptsize$\pm$0.13 & \scriptsize$\pm$0.34 & \scriptsize$\pm$1.24 & \scriptsize$\pm$3.06 & \scriptsize$\pm$1.29 \\
\midrule
\multicolumn{9}{l}{\textit{\small LLaVA-1.5-Qwen2.5-7B}} \\[-2pt]
Baseline & 61.1 & 87.2 & 42.7 & 40.1 & 57.8 & 12.2 & 312.0 & 56.4 \\
\rowcolor{baselinecolor}
\ours
  & \textbf{62.2} \textcolor{softgreen}{(+1.1)}
  & \textbf{87.7} \textcolor{softgreen}{(+0.5)}
  & \textbf{43.2} \textcolor{softgreen}{(+0.5)}
  & \textbf{41.8} \textcolor{softgreen}{(+1.7)}
  & \textbf{58.7} \textcolor{softgreen}{(+0.9)}
  & \textbf{15.3} \textcolor{softgreen}{(+3.1)}
  & 311.3 \textcolor{softgray}{(-0.7)}
  & \textbf{59.0} \textcolor{softgreen}{(+2.6)} \\
\rowcolor{baselinecolor}
  & \scriptsize$\pm$1.53 & \scriptsize$\pm$0.33 & \scriptsize$\pm$0.64 & \scriptsize$\pm$0.75 & \scriptsize$\pm$0.64 & \scriptsize$\pm$2.48 & \scriptsize$\pm$6.35 & \scriptsize$\pm$0.40 \\
\midrule
\multicolumn{9}{l}{\textit{\small OneVision-1.5-4B}} \\[-2pt]
Baseline & 70.0 & 88.6 & 55.3 & 48.8 & 65.7 & 18.69 & 627.0 & 66.1 \\
\rowcolor{baselinecolor}
\ours
  & \textbf{70.9} \textcolor{softgreen}{(+0.9)}
  & \textbf{88.7} \textcolor{softgreen}{(+0.1)}
  & \textbf{55.4} \textcolor{softgreen}{(+0.1)}
  & \textbf{52.5} \textcolor{softgreen}{(+3.7)}
  & \textbf{66.9} \textcolor{softgreen}{(+1.2)}
  & 17.37 \textcolor{softgray}{(-1.32)}
  & \textbf{634.0} \textcolor{softgreen}{(+7.0)}
  & \textbf{66.4} \textcolor{softgreen}{(+0.3)} \\
\rowcolor{baselinecolor}
  & \scriptsize$\pm$0.96 & \scriptsize$\pm$0.33 & \scriptsize$\pm$0.52 & \scriptsize$\pm$0.79 & \scriptsize$\pm$0.25 & \scriptsize$\pm$0.09 & \scriptsize$\pm$3.54 & \scriptsize$\pm$0.09 \\
\bottomrule
\end{tabular}}
\label{tab:main_results}
\vspace{-9pt}
\end{table}
\begin{table}[t]
\centering
\small
\setlength{\tabcolsep}{2.5pt}
\caption{Effect of SSL tasks with LoRA training on LLaVA-1.5-Qwen2.5. Numbers in parentheses indicate improvement over the baseline. Baseline results are reproduced, while VIRAL results are reported from~\cite{yoon2025visual}. POPE$^{*}$ represents average accuracy across the “random” and “popular” subsets.}
\resizebox{0.85\textwidth}{!}{%
\begin{tabular}{llccccc}
\toprule
\textbf{Model} & \textbf{Method} & \textbf{CVB-2D} & \textbf{POPE}$^{*}$ & \textbf{MMStar} & \textbf{BLINK} & \textbf{Avg.} \\
\midrule
\multirow{3}{*}{\shortstack{LLaVA-1.5-Qwen2.5\\ (LoRA)}}
& Baseline & 59.9 & 87.7 & 42.6 & 40.0 & 57.4 \\
& \cellcolor{baselinecolor} \ours
  & \cellcolor{baselinecolor} \textbf{63.8} \textcolor{softgreen}{(+3.9)}
  & \cellcolor{baselinecolor} \textbf{88.5} \textcolor{softgreen}{(+0.8)}
  & \cellcolor{baselinecolor} \textbf{43.7} \textcolor{softgreen}{(+1.1)}
  & \cellcolor{baselinecolor} \textbf{43.3} \textcolor{softgreen}{(+3.3)}
  & \cellcolor{baselinecolor} \textbf{59.6} \textcolor{softgreen}{(+2.2)} \\
& \,\texttt{VIRAL}~\citep{yoon2025visual} & 60.5 & 88.3 & 39.2 & -- & -- \\
\bottomrule
\end{tabular}}
\label{tab:lora_results}
\vspace{-11pt}
\end{table}

In \autoref{tab:lora_results}, we evaluate our method on LLaVA-1.5-Qwen using LoRA fine-tuning~\citep{hu2022lora}, a parameter-efficient adaptation approach, instead of full fine-tuning. The results demonstrate substantial performance improvements even in this parameter-efficient setting. We also compare against VIRAL~\citep{yoon2025visual}, a recent method that enhances vision-centric benchmarks via auxiliary distillation losses (reported in the LoRA setting). Despite its simplicity and the absence of additional objectives or architectural modifications, our approach achieves higher overall performance. These findings suggest that carefully adjusting the instruction tuning distribution can be more effective than introducing specialized loss functions for improving vision-centric instruction-following performance.

We report results on general benchmarks in \autoref{tab:main_results}. We observe improvements on MathVision and RealWorldQA for  LLaVA-1.5-Qwen  and on OCRBench for OneVision-1.5, on-par for  LLaVA-1.5-Vicuna, indicating that injecting vision-centric SSL tasks does not hurt general reasoning skills.


\subsection{Experimental Analysis}
\label{sec:ablations}
\pgfplotsset{compat=1.17}
\begin{figure}[t]
    \centering
    \begin{tikzpicture}
    \begin{axis}[
        width=7.2cm, height=3.5cm,
        ylabel={Avg. Score (\%)},
        symbolic x coords={0,1,3,5,10,30},
        xtick=data,
        xticklabels={{$\rho$=0\%},{1\%},{3\%},{5\%},{10\%},{30\%}},
        ymin=57.4, ymax=59.9,
        ymajorgrids=true,
        grid style={dashed, gray!30},
        mark size=2.5pt,
        title={\small{\textbf{LLaVA-1.5 Qwen-2.5-7B}}},
        ylabel near ticks,
        xlabel near ticks,
        tick label style={font=\small},
        label style={font=\small},
        title style={font=\normalsize, , yshift=-5pt},
        axis line style={gray!60},
    ]
    \addplot[
        color=blue!70,
        mark=*,
        mark options={fill=blue!70, draw=blue!90},
        line width=1.5pt,
    ] coordinates {
        (0,  57.8)
        (1,    58.2)
        (3,    58.1)
        (5,    58.6)
        (10,   58.7)
        (30,   58.7)
    };
    \addplot[
        only marks, mark=star, mark size=4pt, color=red!80,
        mark options={fill=red!80, draw=red!90},
    ] coordinates {(10, 58.7)};
    \node[above=3pt, font=\small\bfseries, text=red!80] at (axis cs:10, 58.7) {58.7};
    \end{axis}
    \end{tikzpicture}
    \hfill
    \begin{tikzpicture}
    \begin{axis}[
        width=7.2cm, height=3.5cm,
        ylabel={Avg. Score (\%)},
        symbolic x coords={0,1,3,5,10,30},
        xtick=data,
        xticklabels={{$\rho$=0\%},{1\%},{3\%},{5\%},{10\%},{30\%}},
        ymin=65.3, ymax=67.6,
        ymajorgrids=true,
        grid style={dashed, gray!30},
        mark size=2.5pt,
        title={\small{\textbf{OneVision-1.5-4B}}},
        ylabel near ticks,
        xlabel near ticks,
        tick label style={font=\small},
        label style={font=\small},
        title style={font=\normalsize, yshift=-5pt},
        axis line style={gray!60},
    ]
    \addplot[
        color=blue!70,
        mark=*,
        mark options={fill=blue!70, draw=blue!90},
        line width=1.5pt,
    ] coordinates {
        (0,  65.7)
        (1,    65.82)
        (3,    66.89)
        (5,    66.19)
        (10,   66.39)
        (30,   66.34)
    };
    \addplot[
        only marks, mark=star, mark size=4pt, color=red!80,
        mark options={fill=red!80, draw=red!90},
    ] coordinates {(3, 66.89)};
    \node[above=3pt, font=\small\bfseries, text=red!80] at (axis cs:3, 66.89) {66.9};
    \end{axis}
    \end{tikzpicture}
    \caption{Effect of the SSL injection ratio $\rho$ on vision-centric instruction-following performance for LLaVA-1.5-Qwen2.5-7B (left) and OneVision-1.5-4B (right).}
    \label{fig:vg_ablation}
\vspace{-10pt}
\end{figure}
\begin{table}[t]
\centering

\begin{minipage}[t]{0.49\linewidth}
\centering
\small
\caption{ \textbf{SSL task ablation.} Impact of individual and combined SSL pretext tasks on LLaVA-1.5-Qwen2.5-7B ($\rho=1\%$ per task). 
}
\label{tab:task_ablation}
{\setlength{\tabcolsep}{4pt}
\resizebox{\linewidth}{!}{%
\begin{tabular}{ccc@{\hspace{8pt}}ccccc}
\toprule
\multicolumn{3}{c@{\hspace{12pt}}}{SSL task} & \multicolumn{5}{c}{Benchmark} \\
\textbf{Rot.} & \textbf{Col.} & \textbf{Corr.} & \textbf{CVB-2D} & \textbf{POPE} & \textbf{MMStar} & \textbf{BLINK} & \textbf{Avg.} \\
\midrule
& & & 61.1 & 87.2 & 42.7 & 40.1 & 57.8 \\
\checkmark & & & \textbf{62.4} & 87.2 & 42.5 & 40.7 & 58.2 \\
& \checkmark & & 62.3 & 87.5 & 41.8 & 41.7 & 58.3 \\
& & \checkmark & 61.9 & 87.3 & 41.7 & \textbf{42.0} & 58.2 \\
\rowcolor{baselinecolor} \checkmark & \checkmark & \checkmark & 62.2 & \textbf{87.7} & \textbf{43.2} & 41.8 & \textbf{58.7} \\
\bottomrule
\end{tabular}}}
\end{minipage}
\hfill
\begin{minipage}[t]{0.49\linewidth}
\centering
\small
\caption{\textbf{Training compute control.} Comparison against baselines trained with the same proportional increase in instruction-tuning iterations, but without SSL tasks, on 
OneVision-1.5-4B.
}
\label{tab:training_compute_impact}
{\setlength{\tabcolsep}{4pt}
\resizebox{\linewidth}{!}{%
\begin{tabular}{l@{\hspace{6pt}}ccccc}
\toprule
\textbf{Training Setup} & \textbf{CVB-2D} & \textbf{POPE} & \textbf{MMStar} & \textbf{BLINK} & \textbf{Avg.} \\
\midrule
Baseline & 70.0 & 88.6 & 55.3 & 48.8 & 65.7 \\

\begin{tabular}[c]{@{}l@{}}Baseline\\[-1pt](+3\% instruct only)\end{tabular}
& 68.2 & \textbf{89.0} & \textbf{55.4} & 49.3 & 65.6 \\

\rowcolor{baselinecolor}
\begin{tabular}[c]{@{}l@{}}\ours\\[-1pt](+3\% SSL)\end{tabular}
& \textbf{70.9} & 88.7 & \textbf{55.4} & \textbf{52.5} & \textbf{66.9} \\
\bottomrule
\end{tabular}}}
\end{minipage}
\end{table}

\parag{Impact of each SSL pretext task.}
In \autoref{tab:task_ablation}, we analyze the effect of incorporating each SSL instruction task individually, as well as their combination, during instruction tuning. 
Each SSL task individually improves average performance over the baseline, and combining all three yields further gains that are both stronger and more consistent across benchmarks. 
This suggests that the tasks provide complementary supervision signals, with their joint use producing the most robust improvements in vision-centric instruction-following.
Rotation prediction and color matching alone yield consistent gains, indicating that improvements are not specific to the correspondence task.


\parag{How much SSL is enough? Impact of injection ratio $\rho$.}
In \autoref{fig:vg_ablation} we evaluate the impact of the SSL injection ratio $\rho$, which controls the proportion of SSL instruction samples added to the instruction tuning dataset, for LLaVA-1.5-Qwen and OneVision-1.5.
Although the two models exhibit different sensitivity to $\rho$, likely due to differences in dataset scale, architecture, and training pipeline, the overall trend is consistent. Even a small fraction of SSL data ($\rho=1\%$) yields measurable improvements over the baseline ($\rho=0\%$). Performance peaks at $\rho=10\%$ for LLaVA-1.5-Qwen and $\rho=3\%$ for OneVision-1.5, after which gains saturate or slightly decline.
These results indicate that modest amounts of SSL-based visually grounded supervision are sufficient to improve vision-centric instruction-following performance, while larger proportions provide limited additional benefit. Since $\rho$ is defined relative to the size of the instruction-tuning dataset, \ours scales the amount of SSL supervision with the instruction tuning data, maintaining a balance between general instruction-following and visually grounded tasks across different training regimes.

\parag{Are Gains Due to Additional Training Compute?}
Injecting SSL tasks with ratio $\rho$ increases the total number of instruction tuning samples, resulting in a proportional increase in training iterations and compute. Although $\rho$ is small in our setting, we verify that the observed improvements do not simply stem from additional training.
In \autoref{tab:training_compute_impact}, we compare our method against a control baseline trained with the same increase in training iterations but using only standard instruction data, without SSL tasks. Specifically, we use the OneVision-1.5 framework and extend training by re-exposing the model to previously seen instruction data.
The results show that additional training compute alone does not improve performance. Gains arise when the extra data consists of SSL tasks, confirming that improvements are driven by the nature of the supervision rather than the slightly longer training.

\begin{table}[t!]
\centering

\begin{minipage}[t]{0.49\linewidth}
\centering
\small
\caption{\textbf{When to inject SSL.} Effect of SSL injection stage on LLaVA-1.5-Qwen2.5-7B. Training after instruction tuning is performed using LoRA to avoid catastrophic forgetting. 
}
{\setlength{\tabcolsep}{4pt}
\resizebox{\linewidth}{!}{%
\begin{tabular}{l@{\hspace{8pt}}ccccc}
\toprule
\textbf{SSL Injection Stage} & \textbf{CVB-2D} & \textbf{POPE} & \textbf{MMStar} & \textbf{BLINK} & \textbf{Avg.} \\
\midrule
Baseline & 61.1 & 87.2 & 42.7 & 40.1 & 57.8 \\
SSL before IT (3\%)& 60.1 & 87.1 & 42.6 & 40.8 & 57.7 \\
SSL before IT (100\%) & 57.6 & 87.6 & 43.9& 39.1 & 57.1 \\
\rowcolor{baselinecolor} SSL during IT (ours) & \textbf{62.2} & \textbf{87.7} & 43.2 & \textbf{41.8} & \textbf{58.7} \\
SSL after IT & 36.5 & 50.2 & 39.5 & 33.2 & 39.9 \\
SSL after IT (GRPO) & 61.1 & 87.4 & \textbf{43.5} & 39.8 & 58.0 \\
\bottomrule
\end{tabular}}}
\label{tab:where_to_inject}
\end{minipage}
\hfill
\begin{minipage}[t]{0.49\linewidth}
\centering
\small
\caption{\textbf{SSL image source.} Effect of SSL image source (rotation prediction, $\rho=1\%$) on LLaVA-1.5-Qwen2.5-7B.
}
{\setlength{\tabcolsep}{4pt}
\resizebox{\linewidth}{!}{%
\begin{tabular}{lccccc}
\toprule
\textbf{SSL Image Source} & \textbf{CVB-2D} & \textbf{POPE} & \textbf{MMStar} & \textbf{BLINK} & \textbf{Avg.} \\
\midrule
Baseline & 61.1 & 87.2 & 42.7 & 40.1 & 57.8 \\
\begin{tabular}[c]{@{}l@{}}Single image\\[-1pt](augmented views)\end{tabular}
& \textbf{62.7} & 86.9 & 42.5 & \textbf{41.5} & \textbf{58.4} \\
\rowcolor{baselinecolor}
\begin{tabular}[c]{@{}l@{}}COCO\\[-1pt](reused images)\end{tabular}
& 62.4 & \textbf{87.2} & 42.5 & 40.7 & 58.2 \\
\bottomrule
\end{tabular}}}
\label{tab:impact_of_data_source}
\end{minipage}
\vspace{-9pt}
\end{table}


\parag{Where Should SSL Be Applied?}
Our method integrates SSL-based visually grounded tasks during the visual instruction tuning stage. In \autoref{tab:where_to_inject}, we evaluate alternative injection strategies to understand whether timing matters. Specifically, we compare three settings: (a) adding an SSL-only stage before the original instruction tuning, (b) mixing SSL tasks during instruction tuning (our method), and (c) adding an SSL-only stage after the original instruction tuning.

We observe that improvements occur only when SSL tasks are injected during instruction tuning. Applying SSL before instruction tuning yields performance comparable to the baseline, even with increased data size (100\%), suggesting that a subsequent instruction tuning phase largely overrides the effect of a standalone SSL stage. Injecting SSL after instruction tuning causes significant performance degradation due to catastrophic forgetting. 
In this case, even careful tuning of the SSL ratio $\rho$, reducing it from 10\% to 1\%, and using LoRA~\citep{hu2022lora}, cannot fully recover general instruction-following ability.
Training with GRPO~\citep{shao2024deepseekmath} avoids this collapse and even slightly improves POPE and MMStar over the baseline, but the gains are not consistent across benchmarks and the average score remains below SSL during instruction tuning. 
This suggests that while RL post-training can preserve general ability better than supervised finetuning, it does not replace the benefit of exposing the model to visually grounded instructions during the main instruction-tuning stage.
Overall, these results indicate that integrating SSL supervision directly within instruction tuning is both more effective and more robust than introducing separate pre- or (RL-based) post-training SSL stages.

\parag{Effect of SSL Image Source.}
In \autoref{tab:impact_of_data_source}, we analyze how the choice of image source for constructing SSL tasks influences vision-centric instruction-following performance, using rotation prediction as a representative SSL objective. We consider two contrasting settings: (i) COCO, which is already part of the original instruction tuning data and can be reused without introducing additional images, and (ii) a single high-resolution image \citep{asano2020critical}, from which multiple augmented views are generated via geometric cropping and color jittering to create diverse SSL samples.
Both sources lead to consistent improvements over the baseline on vision-centric benchmarks. Notably, even SSL supervision derived from multiple views of a single image provides gains. These findings suggest that the critical factor is not dataset scale or diversity, but the presence of visually grounded objectives that force the model to rely on visual input. In such tasks, language priors alone are insufficient, encouraging stronger alignment between visual tokens and the language model.


\begin{figure}[t]
\centering
\setlength{\tabcolsep}{0.3pt}
\begin{tabular}{cccc}
    \small Baseline & \small \textbf{\ours} & \small Baseline & \small \textbf{\ours} \\
    \includegraphics[width=0.23\linewidth, trim={0 0 0 1cm}, clip]{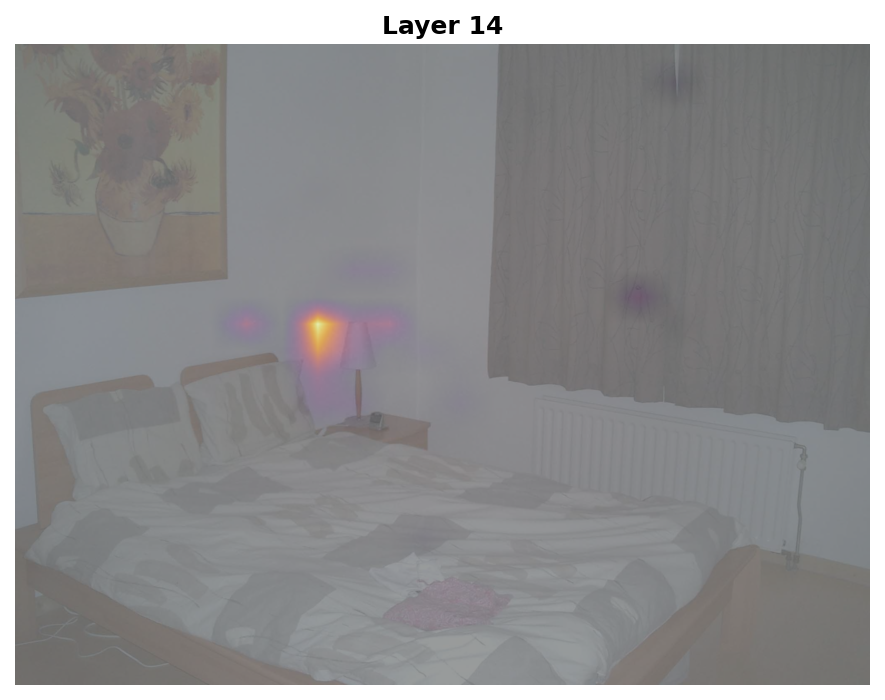} &
    \includegraphics[width=0.23\linewidth, trim={0 0 0 1cm}, clip]{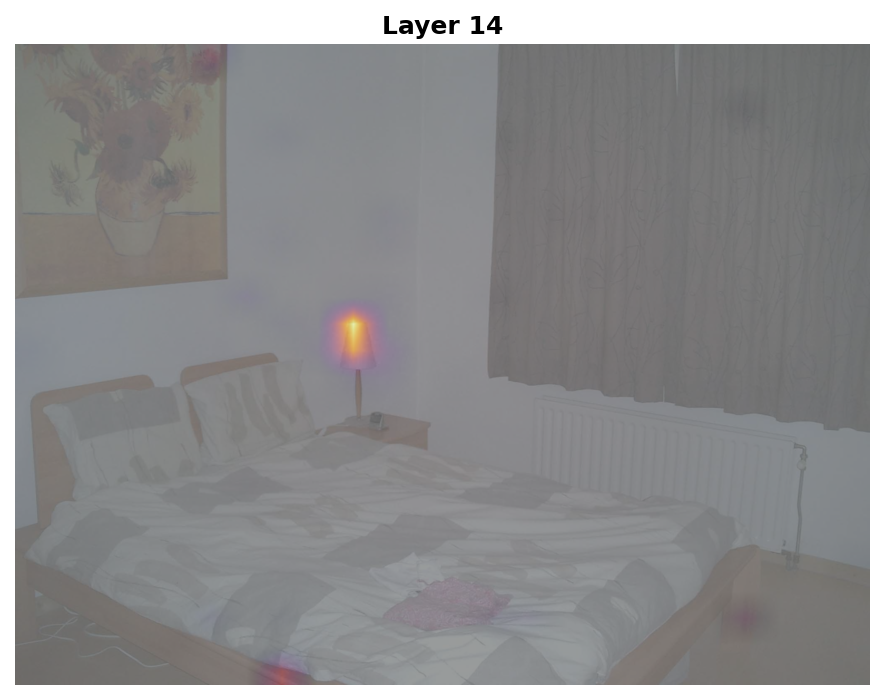} &
    \includegraphics[width=0.23\linewidth, trim={0 0 0 1cm}, clip]{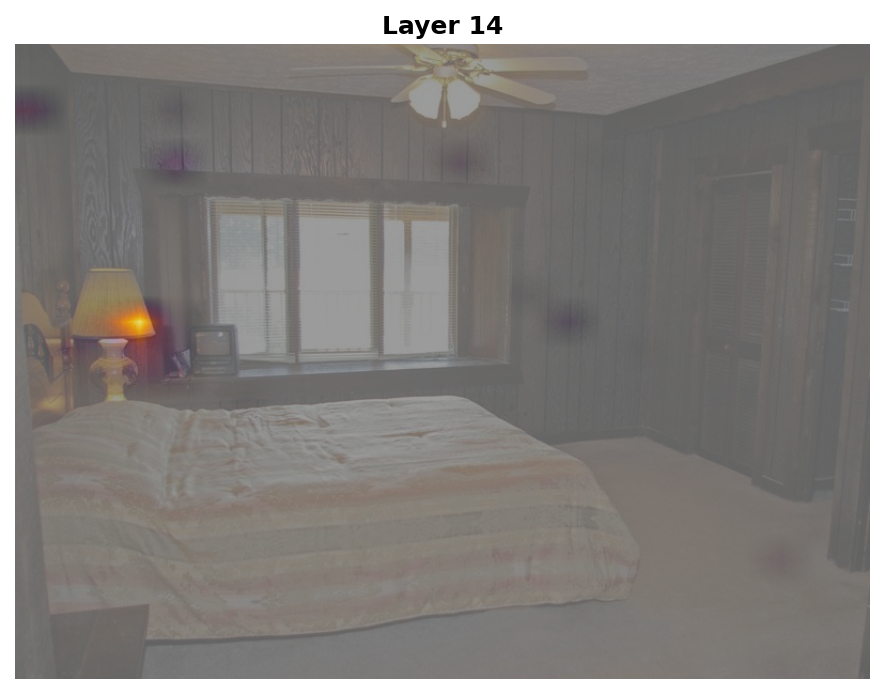} &
    \includegraphics[width=0.23\linewidth, trim={0 0 0 1cm}, clip]{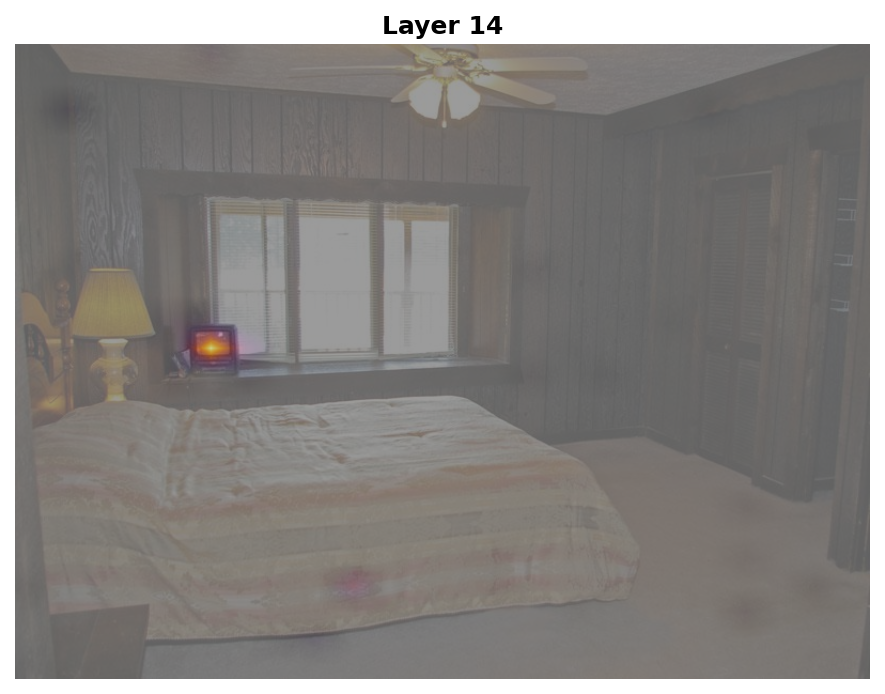} \\
    \multicolumn{2}{c}{\scriptsize Q: How many \textbf{table lamps} are in the image?} &
    \multicolumn{2}{c}{\scriptsize Q: How many \textbf{televisions} are in the image?}  \\
\end{tabular}
\caption{Attention map from the Baseline (LLaVA-1.5-Vicuna-7B) and \ours on CVB-2D examples. \ours produces \emph{more focused and better localized} attention on task-relevant objects.}
\label{fig:attn}
\end{figure}
\begin{figure}[t!]
    \centering
    \input{figures/qualitatives_tikz}
    \caption{\textbf{Qualitative examples.} We present a few qualitative examples comparing LLaVA-1.5 Qwen-2.5-7B baseline against \ours. Our SSL-inspired tasks yield improvements on the variety of vision oriented skills such as counting, multi-view reasoning and visual reasoning.}
    \label{fig:qualitatives}
\end{figure}

\begin{table}[t]
\centering
\small
\begin{minipage}[t]{0.27\linewidth}
\centering
\caption{Mean TVI $\uparrow$ \citep{long2025understanding} on LLaVA-1.5 Vicuna-7B.}
\label{tab:tvi}
\setlength{\tabcolsep}{3pt}
\footnotesize
\begin{tabular}{lcc}
\toprule
\textbf{Method} & \textbf{CVB-2D} & \textbf{MMStar} \\
\midrule
Baseline & 0.1238 & 0.1426 \\
\rowcolor{baselinecolor}
\ours & \textbf{0.1368} & \textbf{0.1430} \\
\bottomrule
\end{tabular}
\end{minipage}\hfill
\begin{minipage}[t]{0.37\linewidth}
\centering
\caption{Linear semantic segmentation probing on PASCAL VOC using visual representations encoded by the LLM. We report mIoU (\%), averaged over three training runs.}
\label{tab:seg_probe}
\setlength{\tabcolsep}{3pt}
\footnotesize
\begin{tabular}{llcc}
\toprule
\textbf{Model} & \textbf{Method} & \textbf{mIoU} & $\boldsymbol{\Delta}$ \\
\midrule
\multirow{2}{*}{\makecell[l]{OneVision-\\1.5-4B}}
& Baseline & 33.9 & -- \\
& \cellcolor{baselinecolor}\ours & \cellcolor{baselinecolor}\textbf{34.1} & \cellcolor{baselinecolor}+0.2 \\
\midrule
\multirow{2}{*}{\makecell[l]{LLaVA-1.5-\\Qwen2.5-7B}}
& Baseline & 44.1 & -- \\
& \cellcolor{baselinecolor}\ours & \cellcolor{baselinecolor}\textbf{44.5} & \cellcolor{baselinecolor}+0.4 \\
\bottomrule
\end{tabular}
\end{minipage}\hfill
\begin{minipage}[t]{0.30\linewidth}
\centering
\caption{VLind-Bench for OneVision-1.5-4B: robustness to commonsense biases and language priors. Results are averaged over three  runs.}
\label{tab:vlind}
\setlength{\tabcolsep}{3pt}
\footnotesize
\begin{tabular}{lcc}
\toprule
\textbf{Method} & \textbf{\makecell{$\mathrm{Acc}_{CB}$}} & \textbf{\makecell{$\mathrm{Acc}_{LP}$}} \\
\midrule
Baseline & 58.2 & 43.0 \\
\rowcolor{baselinecolor}
\ours & \textbf{58.5} & \textbf{43.6} \\
\bottomrule
\end{tabular}
\end{minipage}
\end{table}

\subsection{Analysis of Visual Information Utilization}
\label{sec4:insights}

\parag{\ours~reduces language priors.} We follow a recent language priors quantization technique proposed in~\cite{long2025understanding}. In \autoref{tab:tvi} we present the mean TVI scores on CVBench-2D and MMStar benchmarks comparing baseline LLaVA-1.5 Vicuna 7B trained with standard Instruction Tuning dataset and the model trained with \ours. We observe that model trained with our improved strategy achieves higher TVI scores on both datasets, indicating that the SSL-based auxiliary tasks reduce the model's reliance on language priors.

\paragraph{Semantic information in LLM representations.}
To investigate the effect of \ours on internal visual representations, we evaluate the quality of visual representations encoded by the LLM using a linear semantic segmentation probe on PASCAL VOC \citep{everingham2010pascal}, following~\cite{liu2025visual}. We report the mIoU from the best-performing layer, layer 25 for LLaVA-1.5-Qwen2.5-7B and layer 11 for OneVision-1.5-4B; the same layers were optimal for both \ours and the baseline. We report results averaged over three independent training runs. As shown in \autoref{tab:seg_probe}, \ours yields improvements across both model architectures, despite the absence of segmentation supervision during training. These results suggest that \ours  improves the quality of visual features learned by the LLM beyond task-specific adaptation.

\paragraph{Robustness to language priors.}
We further evaluate reliance on visual information using VLind-Bench \citep{lee2025vlind}, which measures the ability to overcome commonsense biases and language priors. $\mathrm{Acc}_{CB}$ and $\mathrm{Acc}_{LP}$ denote commonsense bias and language prior accuracy, respectively. As shown in \autoref{tab:vlind}, which reports averaged results over three training runs, \ours improves performance for both commonsense biases and language priors. These results complement our TVI analysis, providing further evidence that visually grounded SSL instructions encourage greater reliance on visual evidence rather than language-based shortcuts.

\parag{\ours~improves attention to visual details.}  In \autoref{fig:attn} we visualize attention maps on CVBench-2D, comparing where the baseline LLaVA-1.5-Vicuna and the one trained with \ours focus attentions when processing visual inputs. We notice that the \ours model concentrates attention more precisely on the relevant objects (the lamp and the television). This suggests that SSL-augmented instruction tuning encourages the model to ground its responses in localized visual evidence.

\parag{Qualitative examples on vision-centric benchmarks.}
We present in \autoref{fig:qualitatives} qualitative examples comparing LLaVA-1.5-Qwen (Baseline) against \ours~across visually demanding tasks, including counting, spatial relation understanding, visual reasoning, multi-view reasoning, and functional correspondence. In each case, the baseline fails by defaulting to plausible language-driven responses, while our model produces the correct answer by integrating visual information more effectively.

\section{Conclusion}

This paper introduces \ours, a novel framework designed to leverage self-supervised learning within the visual instruction tuning pipeline. We instantiate \ours using three diverse pretext tasks (image rotation, colorization, and point correspondences) which are integrated seamlessly without requiring architectural modifications nor modifications to training recipes. Our experiments 
show that \ours yields consistent performance gains across a variety of vision-centric MLLM benchmarks while maintaining competitive generic reasoning capabilities and requiring minimal additional computational overhead. Furthermore, we provide qualitative analyses showing that our approach enables models to better ground their reasoning in fine-grained visual information. We hope \ours will stimulate further research into the effective integration of low-level visual cues for complex multi-modal reasoning. Future work will explore extending our approach to other modalities, e.g., 3D point clouds or audio inputs.

\section*{Acknowledgments}
This work was supported by the European Union’s Horizon Europe
research and innovation programme under grant agreement number 101214398 (ELLIOT), by HPC resources from GENCI-IDRIS (Grants AS011017181, AD011015037R2, AD011015037R1), and project RODEO (ANR-24-CE23-5886).

{
\small
\bibliographystyle{plain}
\bibliography{main}
}

\appendix

\clearpage

\appendix

\section{Limitations}\label{sec:limitations}
Our current instantiation of \ours explores three SSL-style tasks: rotation prediction, point-wise color matching, and point correspondence. 
While these tasks provide complementary visually grounded supervision, they cover only a subset of abilities that may benefit MLLMs. 
A broader exploration of SSL-inspired instruction tasks can be beneficial for tasks involving other modalities, 3D point clouds or audio inputs.

Our experiments are conducted on LLaVA-1.5 variants and OneVision-1.5, which provide sufficiently open training recipes and instruction-tuning data for modifying the instruction distribution. 
Extending the study to other strong open-weight MLLMs remains challenging when the corresponding instruction-tuning data are not fully released. 

The optimal SSL injection ratio may need to be adjusted for different model families or data scales. 
The additional computational cost scales linearly with the amount of SSL data added; however in our experiments this overhead is small because we use low injection ratios.

\section{Broader Impacts}
\label{sec:broader_impacts}

\ours aims to improve visual grounding and reduce reliance on language priors in MLLMs, which may benefit applications requiring more reliable visual understanding across domains such as education, healthcare, and communication. 
At the same time, stronger visual reasoning capabilities may also improve systems used in sensitive or harmful downstream applications. It is important to remain mindful of potential risks, downstream use of improved MLLMs should be accompanied by appropriate safeguards, evaluation for bias and robustness, and application-specific risk assessment.

\section{Implementation Details} \label{sec:implementation_details}
\label{app:implementation_details}

\subsection{Self-supervised instruction-tuning tasks}
\label{app:tasks}

\subsubsection{Colorization task.}
We construct a colorization-based visual reasoning task from the COCO 2017 training split \citep{lin2014microsoft}, discarding grayscale images. 
For each image, we sample (N=5) points, each located at least 20 pixels from the image boundary. The color associated with the $i$-th point is defined as the mean RGB value $c_i \in \mathbb{R}^3$ computed over a $r \times r = 5 \times 5$ neighborhood centered at that point.

To avoid ambiguity, we enforce pairwise distinct colors among the five sampled points by requiring a minimum Euclidean distance of $\delta = 40$ in RGB space, i.e., $|c_i - c_j|_2 \ge \delta$ for $i \neq j$, implemented via rejection sampling. Each RGB value is mapped to a human-readable color name using the XKCD color vocabulary\footnote{The XKCD color vocabulary: https://xkcd.com/color/rgb/}, using nearest-neighbor retrieval in RGB space.

The image is then converted to grayscale, and the sampled locations are annotated with labeled markers (e.g., $A, B, \dots, E$), rendered as filled red circles with text labels. The five ground-truth colors are randomly shuffled and presented to the model as a numbered list in the format \texttt{RGB(r, g, b) (color name)}. The model must recover the correspondence between point labels and shuffled colors and output the mapping in the format $\text{``}A\text{-\{}y_A\text{\}},B\text{-\{}y_B\text{\}},\dots\text{''}$, where $y_A$ denotes the index of the true color for point $A$. Example tasks are shown in \autoref{fig:supp_ssl_tasks} (a).

\subsubsection{Point correspondence task.}  
We construct a point correspondence task from a subset of paired images in the COCO 2017 training split using precomputed self-supervised segmentation masks and a fixed list of image pairs from \cite{sirko2025dip}. For details on pseudo-segmentation mask extraction, we refer the reader to \cite{sirko2025dip}.

Given a pair of images \(I_1, I_2 \in \mathbb{R}^{H \times W \times 3}\) and their corresponding pseudo-object segmentation masks \(M_1, M_2 \in \{0,1\}^{H \times W \times K}\), where \(K\) denotes the number of pseudo-classes (obtained in \cite{sirko2025dip} via K-means clustering of DINOv2-ViT-B/14 \citep{darcet2023vision} features), we first select an object of interest. Specifically, for each pseudo-class \(k \in \{1,\dots,K\}\), we compute the union of the corresponding regions across both masks, \(M_1^k \cup M_2^k\), and select the class with the largest pixel area:  

\[
k^* = \underset{k \in \{1,\dots,K\}}{\mathrm{argmax}} \ |M_1^k \cup M_2^k|
\] 
where \(|\cdot|\) denotes pixel cardinality. The selected pseudo-label \(k^*\) defines the object of interest for the pair.

We then extract dense visual features using DINOv2-ViT-B/14 \citep{darcet2023vision}. Specifically, we use the patch tokens from the final layer, yielding feature maps \(A_1, A_2 \in \mathbb{R}^{h \times w \times 768}\) for the two images. A query point \(q\) is sampled uniformly from the selected region $k^*$ in the first image such that $M_1^{k^*}(q)=1$.  Its corresponding patch feature \(A_1(q) \in \mathbb{R}^{768}\) is matched against all patch features in the second image using cosine similarity $\mathrm{sim}(A_1(q), A_2(j)) =  
\frac{A_1(q)\cdot A_2(j)}{||A_1(q)||,||A_2(j)||}$.  
The corresponding point \(q^+\) is defined as the patch within the selected region \(k^*\) in \(I_2\) with the highest similarity:  

\[  
q^+ = \underset{j \in M_2^{k^*}}{\mathrm{argmax}} \ \mathrm{sim}(A_1(q), A_2(j)),  
\]  
where, by abuse of notation, \(j \in M_2^{k^*}\) indicates that the candidate patch \(j\) is restricted to the region corresponding to pseudo-class \(k^*\). The final position of $q^+$ is taken as the center of the selected patch.

To form a 3-way multiple-choice task, we additionally randomly sample two distractor points from the same region \(M_2^{k^*}\) in the second image. The three candidate points are then randomly shuffled and labeled \((0,1,2)\).

For single-image MLLMs, such as LLaVA-1.5, each example is rendered as a side-by-side composite of $I_1$ (left) and $I_2$ (right), with the query point shown in the left image and the candidate points in the right image. For models that accept multiple images, such as OneVision-1.5, $I_1$ and $I_2$ are fed separately. In both cases, points are visualized as red circles with text labels, and the model is asked to identify which candidate point in the second image corresponds to the query point on the shared object in the first image. Example tasks are shown in \autoref{fig:supp_ssl_tasks} (b).

\subsubsection{Rotation prediction task.}  
\begin{figure*}[h!]
    \centering

    \includegraphics[width=0.78\textwidth]{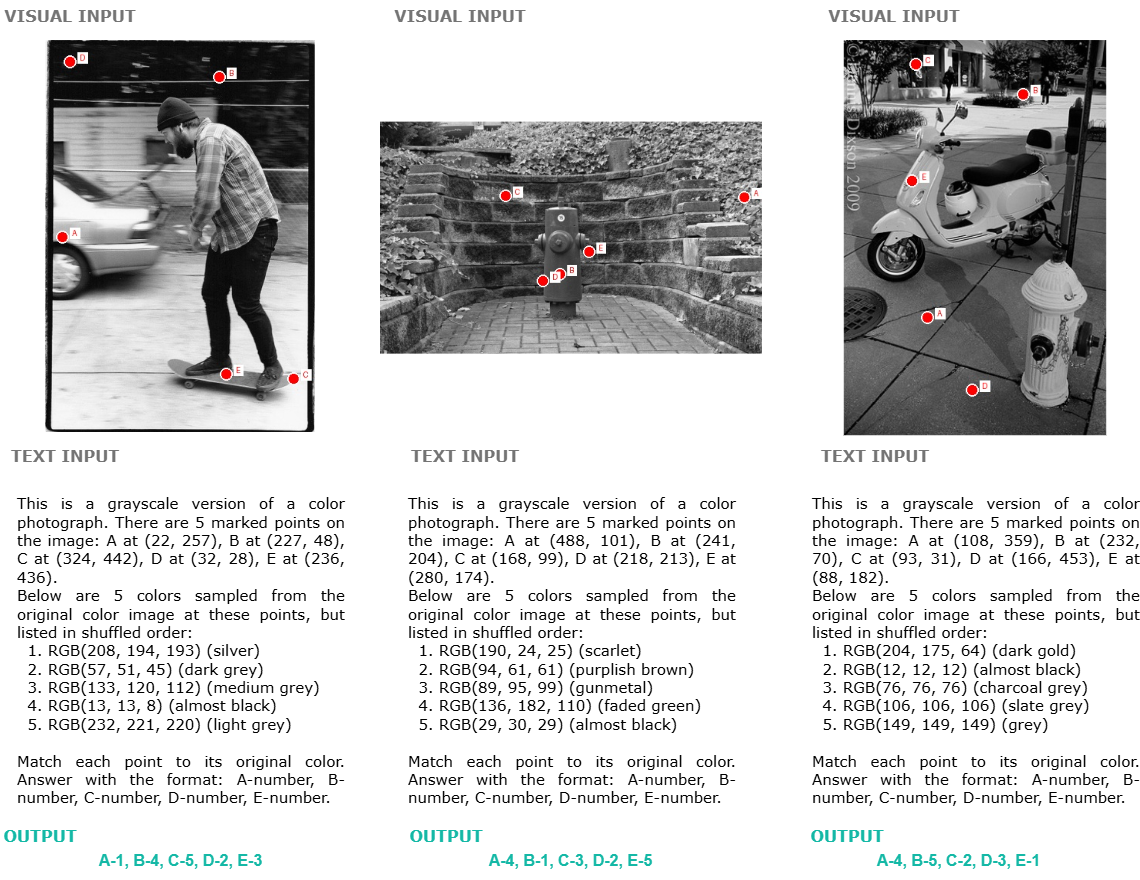}

    \vspace{0.2em}
    {\small (a) Colorization point-matching task.\par}

    \vspace{0.4em}
    \includegraphics[width=0.78\textwidth]{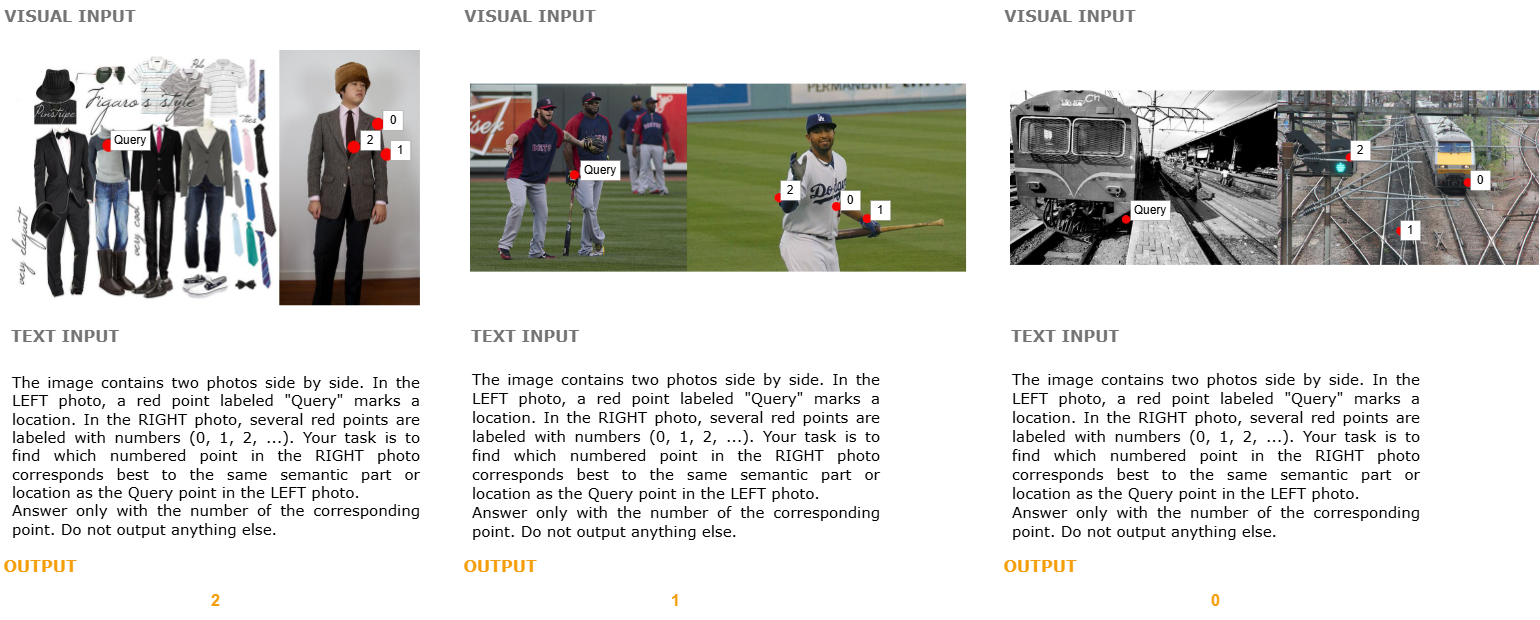}

    \vspace{0.2em}
    {\small (b) Point correspondence task.\par}

    \vspace{0.4em}
    \includegraphics[width=0.78\textwidth]{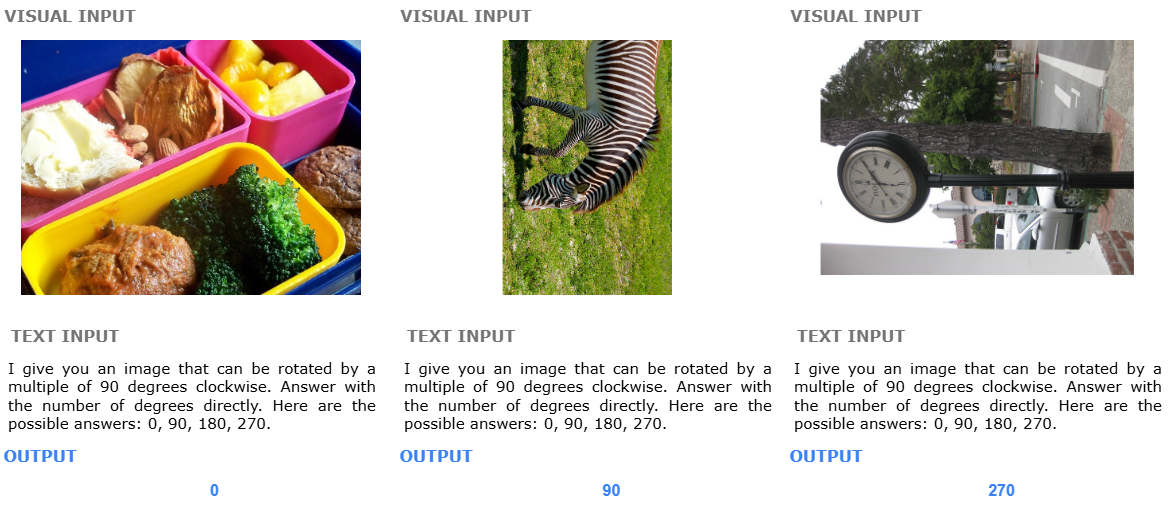}

    \vspace{0.2em}
    {\small (c) Rotation prediction task.\par}

    \caption{Examples of the visually grounded self-supervised tasks used during training: colorization point matching, point correspondence, and rotation prediction.}
    \label{fig:supp_ssl_tasks}
\end{figure*}

We construct a rotation prediction task using images from the COCO 2017 training split \citep{lin2014microsoft}. Each image is rotated clockwise by one of four discrete angles, $\theta \in \{0^\circ, 90^\circ, 180^\circ, 270^\circ\}$. The task is formulated as direct rotation prediction (in degrees): given a single input image, the model receives a fixed prompt stating that the image may be rotated by a multiple of $90^\circ$ clockwise and must respond with one of $\{0, 90, 180, 270\}$. The target output is therefore the integer string corresponding to $\theta$. Example tasks are shown in \autoref{fig:supp_ssl_tasks} (c).

\subsection{Single-image training data construction}

In \autoref{tab:impact_of_data_source} of the main paper (\autoref{sec:ablations}), we evaluate the impact of our method when the images used to construct the self-supervised instruction tuning tasks are derived from a single high-resolution image \citep{asano2020critical}. To obtain visual diversity despite the single-image setting, we generate a large set of augmented views using stochastic cropping and appearance transformations.

Specifically, each sample is produced by applying a random resized crop with crop area uniformly sampled from \([0.1\%, 8\%]\) of the original image area and aspect ratio sampled from \([3/4, 4/3]\), followed by resizing to \(224\times224\) using bilinear interpolation. We then apply horizontal flipping with probability 0.5 and independent random adjustments of brightness, contrast, saturation, and hue. Brightness and contrast factors are sampled uniformly from \([0.75, 1.25]\), saturation from \([0.70, 1.40]\), and hue from \([-0.05, 0.05]\).

\subsection{Evaluation protocols}
All evaluations are conducted using VLMEvalKit~\citep{duan2024vlmevalkit}. We do not rely on external API-based models; instead, we use the toolkit’s built-in answer extraction and matching procedures. We report CVBench-2D~\citep{tong2024cambrian} accuracy, computed over 1,438 2D spatial reasoning questions spanning the Count and Relation sub-tasks; POPE~\citep{li2023evaluating} mean accuracy, computed as the average accuracy across the random, popular, and adversarial COCO-POPE splits; MMStar~\citep{chen2024we}, overall accuracy measured over 1,500 questions spanning six visual reasoning dimensions; and BLINK~\citep{fu2024blink}, overall accuracy measured across 14 visual perception sub-tasks. For general benchmarks, MathVision \citep{wang2024measuring} is reported as accuracy over 3,040 questions spanning 16 mathematical sub-fields, OCRBench \citep{liu2024ocrbench} is reported using the normalized final score, defined as the raw number of correct OCR sub-task answers divided by 10 to obtain a 0--100 scale, and RealWorldQA \citep{xai2024grok} is reported as overall accuracy across all questions. Most results reported in the paper are averaged over three independent training runs.
\begin{table}[t]
\centering
\small
\caption{
Effect of the SSL injection ratio $\rho$ on vision-centric instruction-following performance for OneVision-1.5-4B.}
\setlength{\tabcolsep}{4pt}
\begin{tabular}{r@{\hspace{8pt}}ccccc}
\toprule
\textbf{Injection ratio $\rho$} & \textbf{CVB-2D} & \textbf{POPE} & \textbf{MMStar} & \textbf{BLINK} & \textbf{Avg.} \\
\midrule
1\%  & 69.4 & 88.5 & 55.5 & 49.9 & 65.8 \\
\rowcolor{baselinecolor} 3\%  & \textbf{71.0} & \textbf{88.9} & \textbf{55.5} & 52.2 & \textbf{66.9} \\
5\%  & 69.4 & 88.3 & 54.7 & 52.3 & 66.2 \\
10\% & 69.5 & 88.3 & 54.1 & \textbf{53.7} & 66.4 \\
30\% & 68.9 & 88.2 & 55.3 & 52.9 & 
66.3 \\
\bottomrule
\label{tab:rho_abl}
\end{tabular}
\end{table}

\subsection{Attention map visualization} 

In Figure 4 of the main paper we present attention maps to visual tokens for random samples of CV-Bench2D~\citep{tong2024cambrian}. Specifically, we visualize attention maps of the last token of the input sequence (which corresponds to the last token of the instruction) to all visual tokens in LLaVA-1.5 Vicuna 7B for the baseline and \ours{}. Following \cite{yoon2025visual} we measure the entropy of the attention weights and select the layer with the lowest spatial entropy, hence the strongest focus. We plot an average attention map of this layer through averaging the attention maps of all heads. We notice that \ours{} yields more accurate attentions towards question-relevant objects compared to the baseline model.

\subsection{Training Hyperparameters}
\label{sec:supp_training_hparams}

We provide the main training hyperparameters used for the LLaVA-1.5 and OneVision-1.5 experiments in \autoref{tab:supp_training_hparams} and \autoref{tab:supp_training_hparams_ov}, respectively.
We keep the optimization settings identical to the original instruction-tuning recipes and only modify the instruction-tuning data mixture.

\begin{table}[t]
\centering
\small
\caption{Training hyperparameters for LLaVA-1.5 experiments.}
\setlength{\tabcolsep}{4pt}
\resizebox{\textwidth}{!}{%
\begin{tabular}{lccc}
\toprule
\textbf{Setting} 
& \textbf{LLaVA-1.5-Qwen2.5-7B Full} 
& \textbf{LLaVA-1.5-Vicuna-7B Full} 
& \textbf{LLaVA-1.5-Qwen2.5-7B LoRA} \\
\midrule
LLM backbone & Qwen2.5-7B-Instruct & Vicuna-7B-v1.5 & Qwen2.5-7B-Instruct \\
Vision encoder & CLIP ViT-L/14-336 & CLIP ViT-L/14-336 & CLIP ViT-L/14-336 \\
Training regime & Full fine-tuning & Full fine-tuning & LoRA fine-tuning \\
Epochs & 1 & 1 & 1 \\
Precision & bf16 & bf16 & bf16 \\
Per-device batch size & 1 & 8 & 1 \\
Gradient accumulation steps & 32 & 4 & 32 \\
Number of GPUs & 4 & 4 & 4 \\
Global batch size & 128 & 128 & 128 \\
Learning rate & $2\times10^{-5}$ & $2\times10^{-5}$ & $2\times10^{-4}$ \\
MM projector learning rate & \textemdash & \textemdash & $2\times10^{-5}$ \\
Warmup ratio & 0.03 & 0.03 & 0.03 \\
LR scheduler & cosine & cosine & cosine \\
Max sequence length & 2048 & 2048 & 2048 \\
LoRA rank $r$ & \textemdash & \textemdash & 128 \\
LoRA $\alpha$ & \textemdash & \textemdash & 256 \\
\bottomrule
\end{tabular}}
\label{tab:supp_training_hparams}
\end{table}
\begin{table}[t]
\centering
\small
\caption{Training hyperparameters for OneVision-1.5-4B instruction tuning.}
\setlength{\tabcolsep}{6pt}
\begin{tabular}{lc}
\toprule
\textbf{Setting} & \textbf{OneVision-1.5-4B Full} \\
\midrule
LLM backbone & Qwen3-4B \\
Vision encoder & RICE-ViT, ViT-L/14 \\
Training regime & Full fine-tuning \\
Trainable modules & LLM + adapter + vision encoder \\
Training steps & 3{,}500 \\
Precision & bf16 \\
Global batch size & 224 \\
Learning rate & $1\times10^{-5}$ \\
Min. learning rate & $1\times10^{-6}$ \\
Warmup fraction & 0.002 \\
LR scheduler & cosine \\
Max sequence length & 32{,}768 \\
Max image tokens & 1{,}000 \\
Max image pixels & 2{,}560{,}000 \\
Optimizer & Adam ($\beta_1{=}0.9$, $\beta_2{=}0.99$, $\varepsilon{=}10^{-5}$) \\
Gradient clipping & 1.0 \\
\bottomrule
\end{tabular}
\label{tab:supp_training_hparams_ov}
\end{table}
\section{Additional Results} \label{sec:addition_results}

\paragraph{Detailed results for $\rho$ sensitivity study.}
In \autoref{tab:rho_abl} we present detailed result of the study of $\rho$ parameter which complements \autoref{fig:vg_ablation} of the main paper. In \autoref{tab:rho_general} we present a study of $\rho$ parameter influence on general capabilities. Raising the injection ratio does not systematically degrade general
capabilities. MathVision rises from 12.2 to 19.0 at 30\%; RealWorldQA stays above
baseline at every ratio and reaches 59.1 at 30\%; OCRBench fluctuates around the
baseline with no clear dependence on the ratio.

\begin{table}[t]
\centering
\small
\caption{
Effect of the SSL injection ratio $\rho$ on general capabilities for LLaVA-1.5-Qwen2.5-7B. Results are reported as mean $\pm$ standard deviation over three training runs.}
\setlength{\tabcolsep}{4pt}
\begin{tabular}{r@{\hspace{8pt}}ccc}
\toprule
\textbf{Injection ratio $\rho$} & \textbf{MathVision} & \textbf{OCRBench} & \textbf{RealWorldQA} \\
\midrule
0\% (baseline) & 12.2 $\pm$ 2.1 & 312.0 $\pm$ 4.0 & 56.4 $\pm$ 0.3 \\
1\% & 13.5 $\pm$ 1.2 & 313.3 $\pm$ 4.7 & 58.9 $\pm$ 0.5 \\
3\% & 13.5 $\pm$ 1.2 & \textbf{317.7} $\pm$ 3.8 & 58.0 $\pm$ 0.5 \\
5\% & 15.3 $\pm$ 1.2 & 310.3 $\pm$ 1.2 & 57.0 $\pm$ 0.7 \\
\rowcolor{baselinecolor}
10\% & 15.3 $\pm$ 2.5 & 311.3 $\pm$ 6.4 & 59.0 $\pm$ 0.4 \\
30\% & \textbf{19.0} $\pm$ 1.0 & 313.3 $\pm$ 1.2 & \textbf{59.1} $\pm$ 0.8 \\
\bottomrule
\end{tabular}
\label{tab:rho_general}
\end{table}

\section{Reliability of Self-Supervised Targets}

The self-supervised tasks used in \ours differ in how their training targets are obtained and, consequently, in their susceptibility to label noise. We describe the target generation procedures and assess their reliability below.

\begin{itemize}
    \item \textbf{rotation prediction}. The targets are exact by construction, as the applied rotation angle is sampled from $\{0^\circ, 90^\circ, 180^\circ, 270^\circ\}$.

    \item \textbf{point-wise colorization}. The task is formulated as a multiple-choice matching problem rather than open-ended color prediction. The input image is converted to grayscale, labeled points are overlaid, and the model matches each point to its original color from a shuffled candidate list. To avoid ambiguous candidates, we enforce that the colors sampled at the five points are sufficiently separated: all pairwise Euclidean distances in RGB space must satisfy $\delta \geq 40$. This substantially reduces confusion between visually similar candidate colors. The targets are obtained directly from the original image pixels.

    \item \textbf{point correspondence}.  Unlike the previous tasks, point correspondence relies on pseudo-labels derived from DINOv2 features and pseudo-segmentation masks, which may introduce noise. To quantify the extent of this noise, we conducted a manual audit of 50 correspondence examples drawn uniformly at random from the training set. Five annotators independently assessed whether each provided correspondence target was semantically sensible, and final labels were determined by majority vote. Under this protocol, 41 of the 50 examples (82.0\%) were judged meaningful, suggesting that most pseudo-labels capture plausible semantic correspondences.
    
\end{itemize} 

The single-task ablation in \autoref{tab:task_ablation}, provides a perspective on the contribution of individual tasks. Each task independently improves the average vision-centric score relative to the baseline, including point correspondence despite its imperfect pseudo-labels. The gains from point correspondence also extend to benchmarks beyond correspondence-related tasks. These results show that tasks with different target generation mechanisms and levels of supervision reliability can each contribute to improved visual representations.

\end{document}